%% file: root.tex
\documentclass[letterpaper, 10 pt, conference]{ieeeconf}
\IEEEoverridecommandlockouts
\usepackage{cite}
\usepackage{amsmath,amssymb,amsfonts}
\usepackage{algorithm}
\usepackage{algpseudocode}
\usepackage{algpseudocode}
\usepackage{graphicx}
\usepackage{comment}
\usepackage{url}
\usepackage{hyperref}
\usepackage{textcomp}
\usepackage{xcolor}
\usepackage[nolist]{acronym}
\usepackage{tikz}
\usepackage{siunitx}
\usetikzlibrary{shapes.geometric, arrows.meta, positioning, fit, calc, backgrounds}
\usetikzlibrary{positioning, arrows.meta, fit, backgrounds, calc, shapes.geometric, decorations.markings, patterns, decorations.pathreplacing, fadings, shadows, decorations.pathreplacing, decorations.markings}
\usepackage{booktabs} 
\usepackage{multirow} 
\usepackage{caption}
\usepackage{subcaption}

\definecolor{Blue}{HTML}{0065bd}
\definecolor{Bluelight}{HTML}{2AD5F3}
\definecolor{IRPBlue}{HTML}{03A9F4}
\definecolor{FerrariRed}{HTML}{da1919}

\graphicspath{{./figs/}}

\tikzstyle{block} = [rectangle, rounded corners, minimum width=3cm, minimum height=1cm,text centered, draw=black, fill=blue!20]
\tikzstyle{input} = [ellipse, minimum width=2cm, minimum height=1cm, text centered, draw=black, fill=green!20]
\tikzstyle{output} = [ellipse, minimum width=2cm, minimum height=1cm, text centered, draw=black, fill=red!20]
\tikzstyle{arrow} = [thick,->,>=stealth]

\tikzset{
    basicbox/.style={
        rectangle,
        rounded corners=2pt,
        draw=black!80,
        very thick,
        text centered,
        align=center,
        minimum height=1.2cm,
        minimum width=3.5cm,
        font=\footnotesize
    },
    samplingbox/.style={
        basicbox,
        fill=white
    },
    strategybox/.style={
        basicbox,
        fill=blue!10, 
        draw=blue!80
    },
    costbox/.style={
        basicbox,
        fill=orange!10, 
        draw=orange!80
    },
    arrow/.style={
        -{Latex[length=3mm, width=2mm]},
        thick,
        font=\scriptsize\itshape
    },
    container/.style={
        draw=gray,
        dashed,
        inner sep=0.5cm,
        rounded corners,
        label={[anchor=south, gray, font=\scriptsize]south:#1}
    }
}

\def\BibTeX{{\rm B\kern-.05em{\sc i\kern-.025em b}\kern-.08em
    T\kern-.1667em\lower.7ex\hbox{E}\kern-.125emX}}
\begin{document}


\title{\LARGE \bf A Hybrid Sampling-Based Trajectory Planner\\with Game-Theoretic Guidance for Autonomous Racing}


\author{Alexander Langmann, Frederico Pita de Araujo, Mattia Piccinini and Johannes Betz 
\thanks{A. Langmann, F. Pita, M. Piccinini and J. Betz are with the Professorship of Autonomous Vehicle Systems, TUM School of Engineering and Design, Technical University of Munich, 85748 Garching, Germany; Munich Institute of Robotics and Machine Intelligence (MIRMI), corresponding author: 
\href{mailto:alexander.langmann@tum.de}{alexander.langmann@tum.de}}
}

\maketitle
\begin{abstract}
Autonomous racing demands planning algorithms that balance vehicle dynamics at the limits of handling with strategic decision-making in competitive multi-agent scenarios. Game theory provides a mathematical framework for modeling these interactions, enabling interactive trajectory planning and strategic behaviors, such as blocking. However, directly solving full dynamic games online is computationally prohibitive and challenging to integrate into robust, high-frequency autonomous software stacks.
This paper proposes a hybrid architecture that integrates game-theoretic reasoning into a sampling-based motion planner, combining strategic interactions with robust trajectory generation. Building upon an $\alpha$-potential game formulation, we utilize an offline-learned potential function to capture multi-agent interactions. During online operation, a gradient-based optimization dynamically refines interaction parameters to generate an \textit{Interaction Reference Path}. This path serves as a dynamic cost bias within a high-frequency sampling planner.
We evaluate our approach in a high-fidelity simulation environment on the Yas Marina Circuit. Qualitative and quantitative results demonstrate that our approach successfully induces defensive behaviors like blocking without carrying the computational burden of full dynamic game solvers. 
\end{abstract}

\input{content}

\bibliographystyle{IEEEtran}
\bibliography{literature.bib}

\begin{acronym}
\acro{A2RL}{Abu Dhabi Autonomous Racing League}
\acro{MPC}{Model Predictive Control}
\acro{IRP}{Interaction Reference Path}
\acro{OCP}{Optimal Control Problem}
\acro{IBR}{Iterated Best Response}
\end{acronym}

\end{document}

%% file: content.tex
\section{Introduction}
\label{sec:introduction}
Autonomous racing has demonstrated significant progress in pushing vehicle control and perception to their physical limits \cite{betz2022Autonomous}. However, in multi-vehicle scenarios, autonomous systems still struggle to match the competitive edge of human drivers. This performance gap primarily stems from an insufficient integration of interactive behavior \cite{markkula2020Defining}. Human drivers proactively anticipate and influence their opponents to gain a positional advantage through strategic blocking or overtaking strategies. In contrast, autonomous agents often rely on collision avoidance alone \cite{ogretmen2024SamplingBasedb,piazza2024MPTreea}, which inherently leads to overly conservative behavior (Fig. \ref{fig:RL_intro_a}).

Current motion planning algorithms for handling interactive maneuvers face a fundamental trade-off. Approaches that explicitly model multi-agent interactions, including dynamic game solvers \cite{lecleach2022ALGAMES} or optimal control formulations \cite{rowold2024OpenLoop}, frequently become computationally intractable online due to the severe non-convexity of adversarial racing scenarios. Sampling-based planners \cite{werling2010Optimal, ogretmen2024SamplingBasedb} offer a practical alternative because of their computational robustness and parallelizability. However, standard sampling-based methods treat opponents merely as dynamic obstacles. This reduction leads to purely reactive collision avoidance and lacks any deeper or long-term strategic intent.

\input{figs/teaser_fig}

This paper introduces a hybrid motion planning framework that embeds game-theoretic strategies directly into a sampling-based trajectory planning architecture. Based on the theoretical concept of $\alpha$-potential games \cite{guo2026Markov, guo2025potential}, we use an offline-learned potential function to model complex multi-agent interactions based on \cite{kalaria2025alpharacer}. Instead of solving for a Nash equilibrium within a computationally expensive optimization solver, we apply gradient ascent to the potential function to continuously optimize a set of interaction parameters online. We then use these parameters to construct an \ac{IRP}, which is a geometric reference specifically designed for adversarial maneuvers. The path acts as a strategic bias within the cost function of a sampling-based trajectory planner.

\begin{figure*}[t]
    \centering
    \vspace{1mm}
    \includegraphics[width=0.95\linewidth]{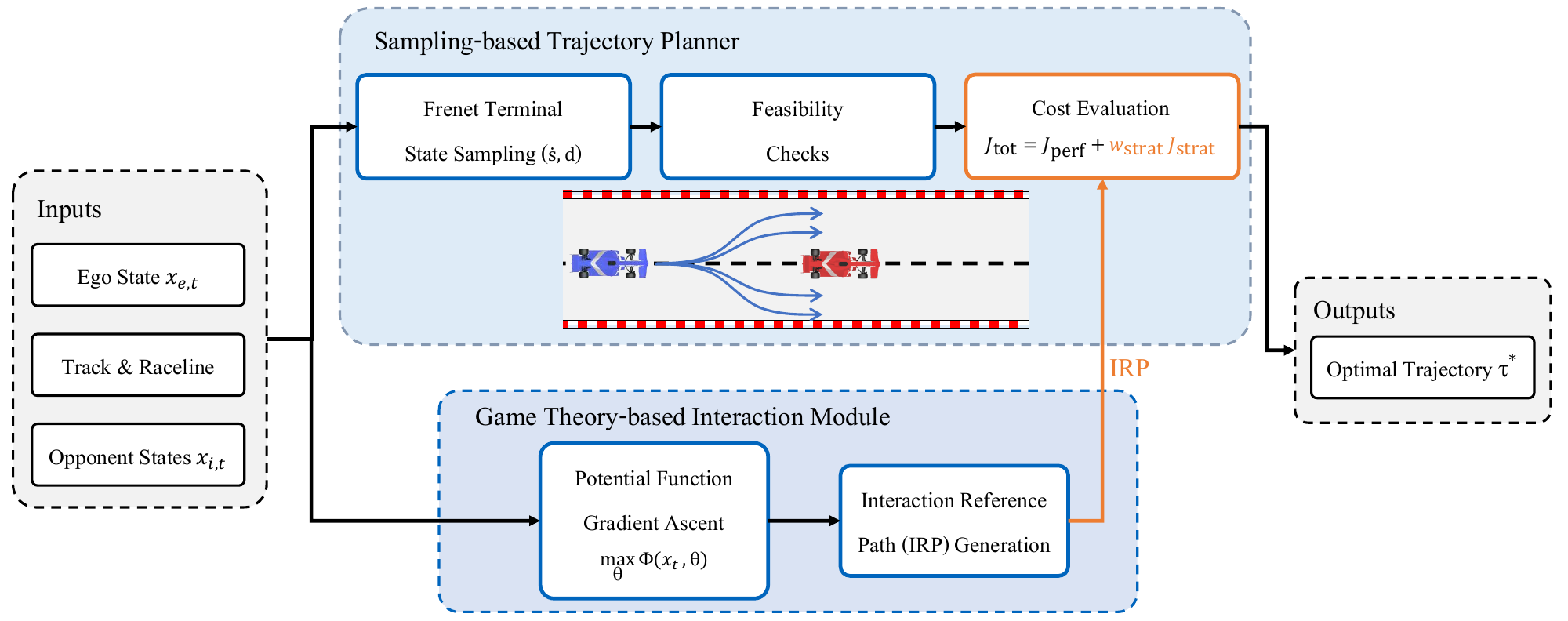}
    \caption{Overview of our proposed method. The framework combines a sampling-based planner based on \cite{ogretmen2024SamplingBasedb} with a game-theoretic interaction module based on \cite{kalaria2025alpharacer}, which generates an \ac{IRP} online. The IRP serves as a strategic cost bias ($J_{\mathrm{strat}}$) during the evaluation of feasible candidate trajectories.}
\label{fig:architecture}
\end{figure*}

The main benefit of this architecture is the decoupling of high-level strategic decisions from trajectory generation. The game-theoretic module dictates the spatial intent, such as committing to a defensive line. Simultaneously, the sampling planner guarantees kinematic feasibility and robust collision avoidance. We demonstrate the validity of this approach using a high-fidelity simulation of the Yas Marina Circuit. The results show emergent defensive behaviors like blocking that standard reactive planners cannot reproduce (Fig. \ref{fig:RL_intro}).

In this work, we present the following key contributions:
\begin{itemize}
    \item A hybrid planning architecture that integrates game-theoretic interaction models into a sampling-based motion planner.
    \item The formulation of an \ac{IRP} generated through online gradient ascent, which serves as a geometric guide for competitive maneuvers.
    \item A systematic simulation study demonstrating emergent blocking behaviors and quantifying the trade-off between strategic positioning and lap time.
\end{itemize}

\section{Related Work}
Motion planning algorithms in autonomous racing must safely navigate multi-vehicle environments while maximizing competitive performance \cite{betz2022Autonomous}. Existing approaches commonly rely on classical architectures, including graph-based, optimization-based, and sampling-based methods. Graph-based planners determine trajectories via interconnected node networks \cite{erke2020improved, rowold2022Efficient, stahl2019Multilayer, ogretmen2022Smoothb}, but struggle with the trade-off between grid resolution and computational feasibility at high speeds. Optimization-based frameworks formulate trajectory generation as an \ac{OCP} over a receding horizon \cite{piccinini2025KinetoDynamical, vazquez2020OptimizationBased, taddei2025Biasing, rowold2023Onlinea, liniger2015Optimizationbased, piccinini2025How}, explicitly incorporating non-linear dynamics and dynamic obstacles \cite{toschi2025Modular}. However, solving non-convex problems online frequently leads to intractability or convergence issues in complex scenarios. Sampling-based planners evaluate dynamically feasible candidate trajectories against a cost function \cite{werling2010Optimal, ogretmen2024SamplingBasedb, langmann2025Online, raji2022Motiona, Langmann2026_multi_stage}. While highly parallelizable and robust against local minima, they suffer from discretization dependencies and computational overhead from evaluating infeasible candidates. All these classical planners natively treat opponents strictly as dynamic obstacles. This restriction forces the ego vehicle into reactive, collision-avoiding behaviors and prevents interactive, strategic maneuvers.

To enable interactive planning \cite{markkula2020Defining}, game theory is applied in urban traffic scenarios for modeling multi-agent interdependencies \cite{fridovich-keil2020Efficient, lecleach2022ALGAMES, schwarting2019Social}. In adversarial racing contexts, prior works have modeled interactions as sequential leader-follower games \cite{liniger2020Noncooperative, jung2025Autonomous, li2023Stackelberg}, or utilized \ac{IBR} \cite{wang2021GameTheoretic, notomista2020Enhancing} for overtaking and blocking. However, finding full dynamic game equilibria in continuous, high-dimensional state spaces is computationally prohibitive, forcing a reliance on simplified kinematic models or unacceptably slow update rates. 

While pure learning-based methods, such as reinforcement learning, can capture complex interactions \cite{evans2023comparing} and achieve remarkable success in car racing simulation \cite{wurman2022Outracinga} or controlled environments for drone racing \cite{kaufmann2023Championlevel}, they require extensive training. Furthermore, physical deployment for full-scale racecars is hindered by safety concerns and the sim-to-real gap. Hybrid architectures address this by embedding learned models into deterministic frameworks, such as using neural networks to shape game-theoretic optimizers \cite{lucente2024DeepGameTP}. Kalaria et al. \cite{kalaria2025alpharacer} approximate non-cooperative dynamic games via an offline-learned $\alpha$-potential function that is maximized during runtime. However, their approach heavily relies on a \ac{MPC} framework, enforcing the game-theoretic output directly as a reference trajectory constraint. 

From the literature review, we deduce a research gap in the form of a robust, sampling-based planner that is informed by game theory to perform truly interactive, rather than purely reactive, maneuvers while remaining online-capable for real-world use.

\section{Methodology}

This section details the proposed hybrid trajectory planning framework. We describe the integration of a game-theoretic reasoning module into a sampling-based planner, the offline learning of the interaction model, the online generation of the \ac{IRP}, and its incorporation into the planning cost function. Figure \ref{fig:architecture} provides an overview of our approach.

\subsection{Hybrid Planning Architecture}
\label{sec:hybrid_framework}
The core trajectory generation layer utilizes a sampling-based local planner based on \cite{ogretmen2024SamplingBasedb}. The planner operates in the Frenet reference frame $(s, d)$, where $s$ denotes the longitudinal progress along a reference trajectory $\Gamma$ and $d$ the lateral offset. As inputs, the planner receives the ego vehicle state $\mathbf{x}_e(t)$, an offline-optimized raceline that serves as the base reference trajectory $\Gamma$, information on race track geometry, and the states and predictions of opponents $\mathbf{x}_i(t)$.

To obtain a trajectory, the planner generates a discrete set of longitudinal profiles $s(t)$ using quartic polynomials by sampling target end velocities $\dot{s}_{\mathrm{end}}$. Concurrently, a set of lateral profiles $d(t)$ is generated using quintic polynomials by varying target lateral states $n_{\mathrm{end}}$, with the motion primitives of \cite{werling2010Optimal}. The longitudinal and lateral profiles are combined to obtain a pool of candidate trajectories $\mathcal{T}$ (Fig. \ref{fig:architecture}). These trajectories are checked for feasibility in terms of compliance with maximum curvature, track bounds, and acceleration limits.
The remaining candidate trajectories $\tau \in \mathcal{T}$ are evaluated against a baseline performance cost function $J_{\mathrm{perf}}(\tau)$. This cost is a weighted sum of metrics penalizing deviations from the target velocity ($c_{\mathrm{\Delta v}}$) and raceline curvature ($c_{\mathrm{\Delta \kappa}}$), lateral offsets from the global raceline ($c_{\mathrm{\Delta rl}}$), and close proximity to an opponent $i$ ($c_{\mathrm{pr}, i}$):
\begin{equation}
\begin{split}
    J_{\mathrm{perf}}(\tau) &= w_{\mathrm{\Delta v}} c_{\mathrm{\Delta v}} + w_{\mathrm{\Delta \kappa}} c_{\mathrm{\Delta \kappa}} \\
    &\quad + w_{\mathrm{\Delta rl}} c_{\mathrm{\Delta rl}} + \sum_{i=1}^{N}w_{\mathrm{pr}, i}c_{\mathrm{pr}, i}
\end{split}
\end{equation}
where $N$ is the number of opponents included in the current planning step.

To enable strategic behaviors such as blocking or overtaking, we augment this reactive architecture with a high-level game-theoretic module based on the work in \cite{kalaria2025alpharacer}. The game-theoretic module computes an \ac{IRP}, which is a geometrical guideline encoding the approximate Nash-equilibrium strategy. This path serves as an additional dynamic bias within the sampling planner’s cost function, to guide the vehicle toward strategic positions, as explained in Section \ref{sec:cost_fun_tot}.

\subsection{Interaction Modeling via $\alpha$-Potential Games}
To capture multi-agent interactions online, we leverage the $\alpha$-potential game formulation of $\alpha$-RACER \cite{kalaria2025alpharacer}. This framework approximates a non-cooperative dynamic game by utilizing a single shared potential function $\Phi(\cdot)$ to capture the strategic incentives of all agents. While we adopt the core mathematical framework, we introduce modifications to the training data generation and the optimized parameter space to suit our decoupled, sampling-based architecture.

\textbf{Value Function Approximation:} Our framework employs a two-stage learning process. First, individual value functions $V_i(x)$ are learned for each participating agent to estimate the expected future progress over a fixed horizon given the current joint state $x$. A key difference from the baseline \cite{kalaria2025alpharacer} lies in our data generation process. Since our standard sampling-based planner uses soft constraints that can implicitly allow physical overlap, we introduce an explicit collision handling mechanism during the generation of training data. Colliding vehicles receive severe velocity and progress penalties. This explicitly encodes a high collision cost into the learned value functions, ensuring the models optimize for safe positional advantages. We learn $V_i(x)$ with a neural network, trained to minimize the Mean Squared Error (MSE) between the predicted progress and the penalized ground truth displacement, using stochastic gradient descent (SGD).

\textbf{Potential Function Training:} In the second stage, the global potential function $\Phi(\mathbf{x}, \boldsymbol{\theta})$ is learned as another neural network, which captures the joint strategic incentives of the interacting agents. The network weights are warm-started by concatenating the pre-trained individual value functions. The training objective minimizes the discrepancy between the gradient of the potential function and the individual value gradients, ensuring that maximizing $\Phi$ approximates a multi-agent Nash equilibrium. Since our underlying sampling-based framework inherently guarantees kinematic feasibility through its sample generation step, our formulation explicitly maps the joint state $\mathbf{x}$ to a reduced, 4-dimensional strategy vector $\boldsymbol{\theta}$ (Section \ref{sec:online_inference}) strictly designed to govern the spatial intent of the ego vehicle. This completely eliminates the need to optimize additional dynamic tracking weights or rigid trajectory constraints within the game-theoretic module.

\subsection{Online Inference and IRP Generation} \label{sec:online_inference}
During online operation, the interaction module determines the optimal strategy parameters $\theta^*$ for the ego vehicle at each time step $t$. Instead of solving a computationally expensive game, we perform a gradient-ascent optimization on the learned potential function:
\begin{equation}
    \boldsymbol{\theta}^* = \arg \max_{\boldsymbol{\theta}} \; \Phi(\mathbf{x}_t, \boldsymbol{\theta}).
\end{equation}
This optimization is executed via backpropagation through the neural network $\Phi(\cdot)$ for a fixed budget of 20 iterations to maintain online feasibility. 
The optimized strategy vector consists of four physically interpretable parameters: $\boldsymbol{\theta} = (p_{\mathrm{vel}}, p_{\mathrm{amp}}, p_{\mathrm{decay}}, p_{\mathrm{block}})$. Here, $p_{\mathrm{vel}}$ scales the target velocity profile, $p_{\mathrm{amp}}$ defines the maximum magnitude of lateral deviation, $p_{\mathrm{decay}}$ governs the spatial decay of the interaction effect, and $p_{\mathrm{block}}$ modulates the blocking aggressiveness based on relative velocity.

These parameters are subsequently used to construct the \ac{IRP}, which serves exclusively as a unified geometric reference for both defensive blocking and active overtaking maneuvers. To construct this path, we adapt the lateral shift functions introduced in \cite{kalaria2025alpharacer} using our parameterization to laterally perturb the offline-calculated optimal raceline $\Gamma$. 

For the ego vehicle executing a block against a relevant opponent $i$, the lateral shift profile $d_{\mathrm{bl},i}(s)$ (blocking component) evaluated at longitudinal coordinate $s$ is calculated based on the opponent's current lateral position $d_i$, the relative velocity $\Delta v_i$, and the spatial distance $(s - s_i)$ to the opponent's longitudinal position $s_i$:
\begin{equation}
    d_{\mathrm{bl},i}(s) = \mathbb{I}_{\mathrm{block}} \cdot (d_i) \left(1 - e^{-p_{\mathrm{block}} \Delta v_i}\right) e^{-p_{\mathrm{decay}} (s - s_i)^2}
\end{equation}
where $\mathbb{I}_{\mathrm{block}}$ is an indicator function activating the block when the opponent approaches from behind. Because the Frenet frame is anchored to the optimal raceline $\Gamma$, the required lateral shift to block the opponent is directly proportional to the opponent's lateral offset $d_i$. For active overtaking maneuvers, the corresponding lateral shift profile $d_{\mathrm{ot},i}(s)$ is calculated in an analogous mathematical manner, utilizing the amplitude parameter $p_{\mathrm{amp}}$ instead of the blocking aggressiveness $p_{\mathrm{block}}$ to determine the lateral deviation. 

The final \ac{IRP} lateral profile $d_{\mathrm{IRP}}(s)$ is the spatial superposition of the base raceline (where $d \equiv 0$) and the interaction-induced lateral shifts of all relevant opponents $i$. We first define the unconstrained profile $d_{\mathrm{raw}}(s)$ as:
\begin{equation}
    d_{\mathrm{raw}}(s) = \sum_i \left(d_{\mathrm{ot},i}(s) + d_{\mathrm{bl},i}(s)\right).
\end{equation}
Because the track width is generally asymmetric relative to the reference line, this raw path is then strictly clipped to the local left and right track boundaries $w_{\mathrm{l}}(s)$ and $-w_{\mathrm{r}}(s)$:
\begin{equation}
    d_{\mathrm{IRP}}(s) = \text{clip}\Big(d_{\mathrm{raw}}(s), \, [-w_{\mathrm{r}}(s), w_{\mathrm{l}}(s)]\Big).
\end{equation}

\subsection{Cost Function Integration}
\label{sec:cost_fun_tot}
The integration of the game-theoretic module into the sampling-based framework is achieved via a composite cost function. Rather than applying rigid trajectory constraints, the planner's baseline performance cost $J_{\mathrm{perf}}(\tau)$ (as defined in Section \ref{sec:hybrid_framework}) is augmented with strategic cost terms derived from the optimized parameter vector $\theta$.

First, the velocity parameter $p_{\mathrm{vel}}$ scales the nominal target velocity $v_{\mathrm{nom}}$ of the global raceline. Consequently, the velocity tracking penalty $c_{\mathrm{\Delta v}}$ within the baseline cost $J_{\mathrm{perf}}(\tau)$ is adjusted to evaluate the candidate trajectory's deviation from this strategically scaled target:
\begin{equation}
    c_{\mathrm{\Delta v}}(\tau) = \int \left(v_{\tau}(t) - p_{\mathrm{vel}} \cdot v_{\mathrm{nom}}(t)\right)^2 dt.
\end{equation}

Second, to embed the lateral strategic intent, we define an additional strategy cost $J_{\mathrm{strat}}(\tau)$. This cost penalizes the lateral deviation between the candidate trajectory $d_{\tau}(s)$ and the dynamically generated \ac{IRP} $d_{\mathrm{IRP}}(s)$. It is formulated as the integral of the absolute lateral error along the longitudinal trajectory path $s$:
\begin{equation}
    J_{\mathrm{strat}}(\tau) = \int \big| d_{\tau}(s) - d_{\mathrm{IRP}}(s) \big| \, ds.
    \label{eq:strategy_cost}
\end{equation}

The total cost used for selecting the optimal trajectory $\tau^*$ is the weighted sum of the modified baseline performance and the lateral strategy cost:
\begin{equation}
    J_{\mathrm{total}}(\tau) = J_{\mathrm{perf}}(\tau) + w_{\mathrm{strat}} \cdot J_{\mathrm{strat}}(\tau)
\end{equation}
where $w_{\mathrm{strat}}$ is a tunable weight that controls the spatial bias towards the \ac{IRP}. This formulation effectively decouples strategic reasoning from safety. The game-theoretic module guides the planner laterally via $J_{\mathrm{strat}}(\tau)$ and longitudinally via $p_{\mathrm{vel}}$, while the hard constraints of the sampling-based trajectory generation and the baseline penalizations in $J_{\mathrm{perf}}(\tau)$ guarantee that the executed maneuver remains dynamically feasible.

\begin{figure*}[h]
    \centering
    \includegraphics[width=0.98\linewidth]{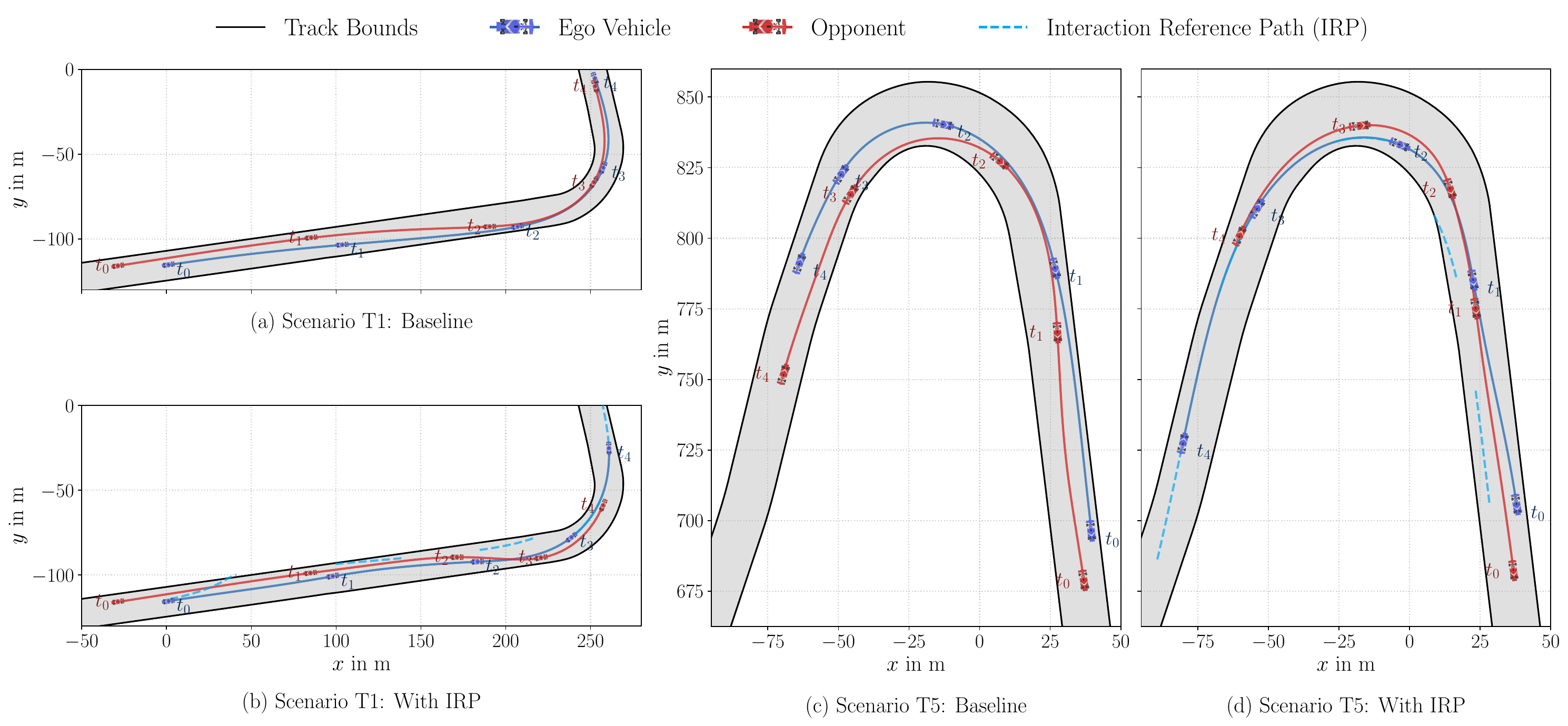}
    \caption{Qualitative comparison of the baseline reactive sampling-based planner (Subfigures a \& c) and our proposed approach (Subfigures b \& d) in scenarios in turn 1 (T1) and turn 5 (T5). Time steps are included at each vehicle position.}
    \label{fig:comparison}
\end{figure*}

\section{Experimental Setup}
\label{sec:results}

\subsection{Simulation Environment}
The proposed framework is evaluated in a high-fidelity multi-agent simulation environment based on a state-of-the-art autonomous racing software stack. The simulation utilizes the track layout of the Yas Marina Circuit. To simulate competitive racing, we introduce one or two opponent vehicles alongside the ego vehicle, respectively. The opponents use the baseline state-of-the-art sampling-based planner \cite{ogretmen2024SamplingBasedb} without game-theoretic interaction capabilities, representing non-cooperative but rational racing agents.

\subsection{Dataset Generation}
To train the interaction model, we generated a dataset consisting of 2247 simulated racing episodes. In each episode, three agents were initialized with random interaction parameters drawn from the ranges specified in Table \ref{tab:ranges}, ensuring a diverse mix of aggressive and passive behaviors.
The simulation utilized a specialized collision handling mechanism during data generation: colliding vehicles received velocity penalties to explicitly encode the cost of crashes into the learned value functions.
\begin{table}[htbp]
    \centering
    \captionsetup{font=footnotesize, textfont=sc, labelfont=sc, labelsep=newline, justification=centering}
    \caption{Overview of interaction parameters and their ranges}
    \label{tab:ranges}
    \small 
    \begin{tabular}{lll}
        \toprule
        \textbf{Parameter} & \textbf{Description} & \textbf{Range} \\
        \midrule
        $p_{\mathrm{vel}}$ & Target velocity scaling & $[0.85, 1.1]$ \\
        $p_{\mathrm{amp}}$ & Lateral shift amplitude & $[0.1, 1.2]$ \\
        $p_{\mathrm{decay}}$ & Spatial decay factor & $[0.01, 0.5]$ \\
        $p_{\mathrm{block}}$ & Blocking aggressiveness & $[0.0, 1.0]$ \\
        \bottomrule
    \end{tabular}
\end{table}

\subsection{Evaluation Scenarios}
\label{sec:scenarios}
To comprehensively evaluate the proposed framework, we define two sets of testing scenarios.
For the qualitative scenarios, we focus on two specific tests to show the planner's interactive capabilities against a single opponent:
\begin{itemize}
    \item \textbf{Straight Line Defense (T1):} The opponent approaches the ego vehicle with a significant speed advantage ($50\,\mathrm{m/s}$ vs. $45\,\mathrm{m/s}$) on a straight leading into turn 1.
    \item \textbf{Hairpin Defense (T5)}: The opponent attempts an overtaking maneuver entering the hairpin at turn 5.
\end{itemize}
In both scenarios, the ego vehicle starts ahead of the opponent on the optimal raceline. We qualitatively compare the emergent behavior of our hybrid planner against the purely reactive baseline planner.
For the quantitative analysis, we define a large-scale randomized scenario set. This encompasses 200 simulation episodes featuring the ego vehicle interacting with two opponent vehicles. The episodes initialize the agents with varying starting velocities in various track locations and randomized grid positions, placing the ego vehicle in either 1st, 2nd, or 3rd position. Both opponents operate using the baseline reactive planner \cite{ogretmen2024SamplingBasedb}.

\section{Results \& Discussion}

\subsection{Training Performance}
\begin{figure}[htbp]
    \centering
    \begin{subfigure}{\linewidth}
        \centering
        \includegraphics[width=\linewidth]{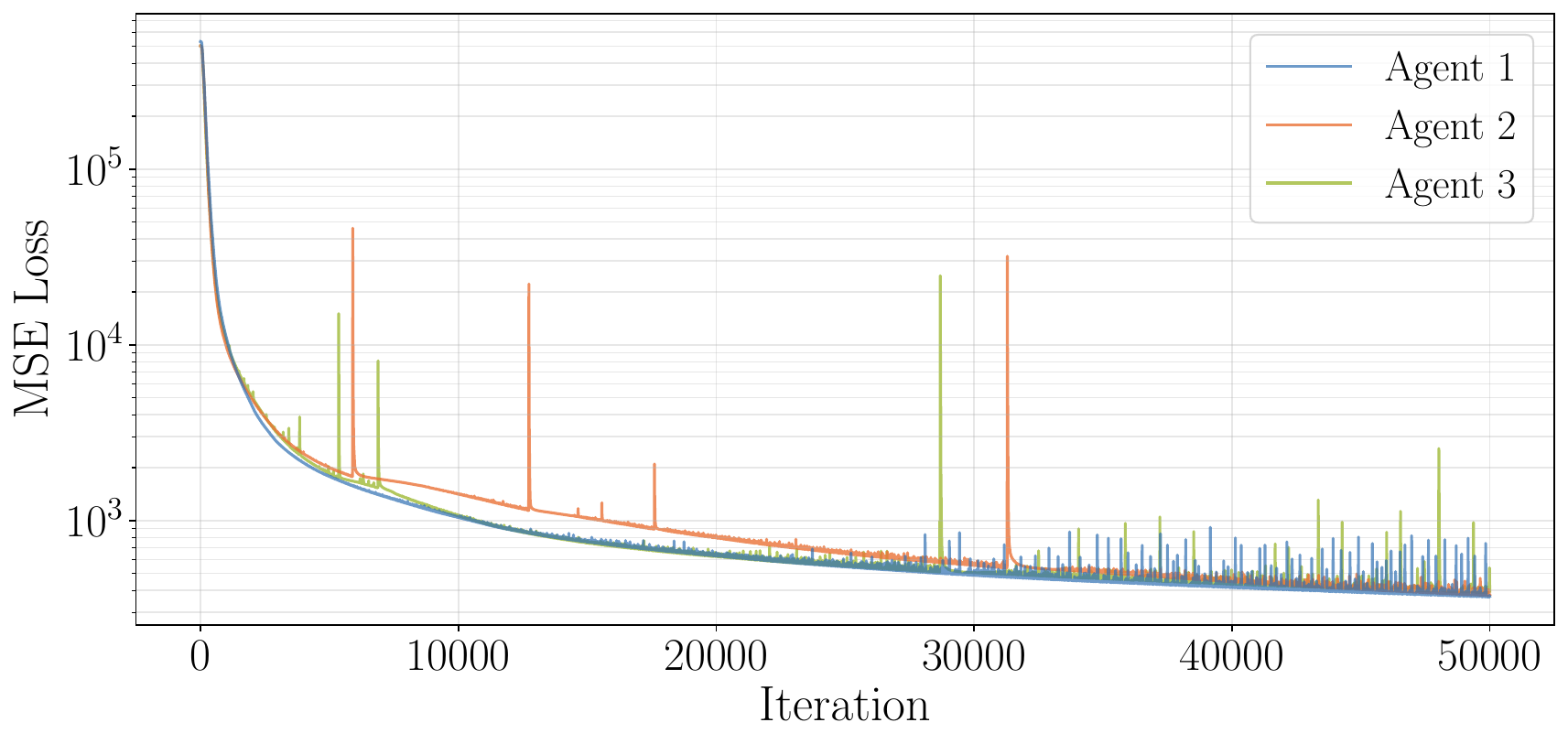}
        \caption{Loss curve for the value functions training.}
        \label{fig:val_func_training}
    \end{subfigure}
    
    \vspace{0.5cm} 
    \begin{subfigure}{\linewidth}
        \centering
        \includegraphics[width=\linewidth]{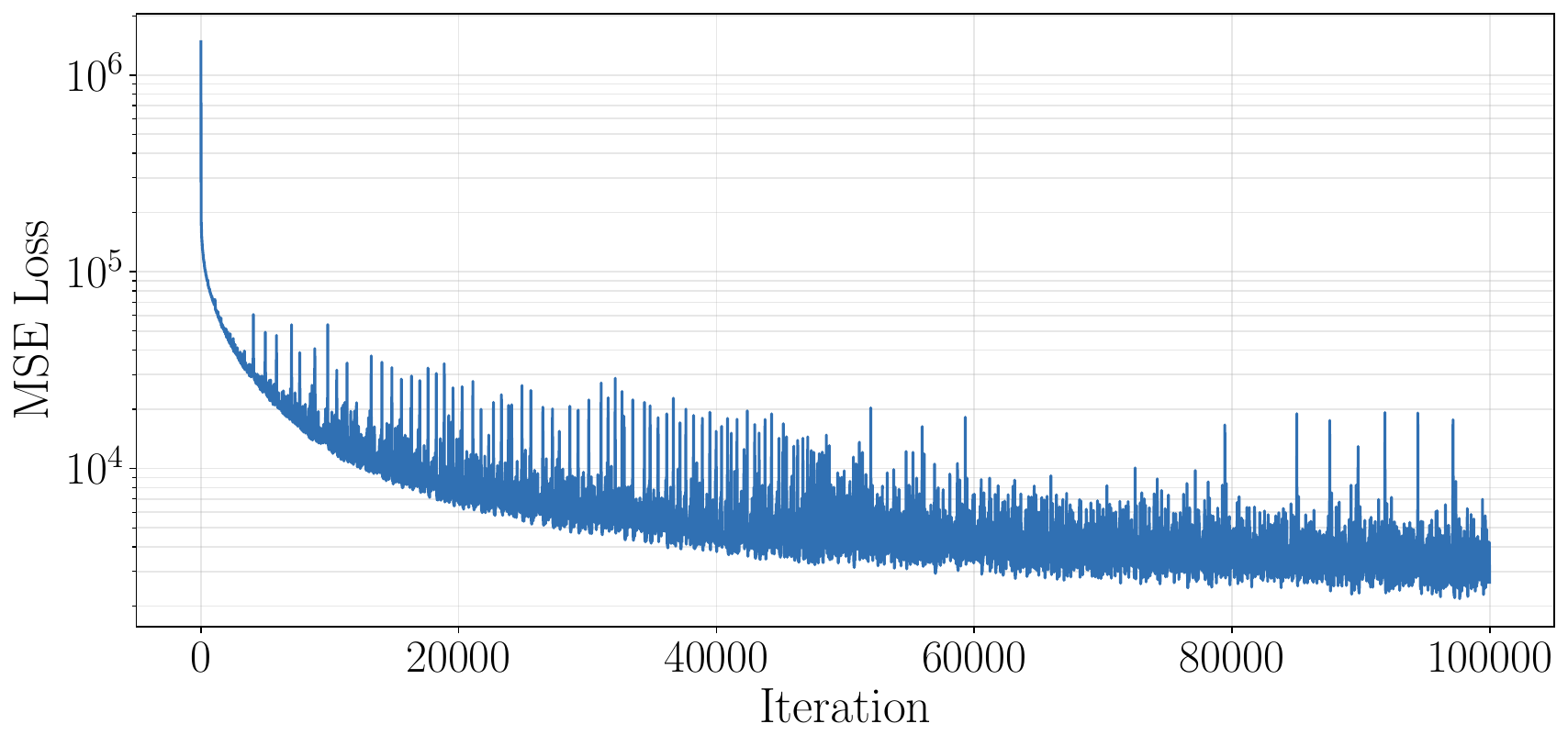}
        \caption{Loss curve for the potential function training.}
        \label{fig:pot_func_training}
    \end{subfigure}
    
    \caption{Training performance of the interaction module. The value functions (a) and the global potential function (b) converge reliably over 50,000 iterations.}
    \label{fig:loss_curves}
\end{figure}

The fidelity of the interaction module depends on the convergence of the underlying neural networks. Figure \ref{fig:loss_curves} shows the loss curves of the value and potential function training.

\textbf{Value Function Convergence}:
The training of individual value functions demonstrated rapid convergence. The MSE loss decreased by 99.92\% over 50,000 iterations. This indicates that the network successfully learned to map vehicle states to expected future progress, a prerequisite for meaningful potential function training.

\textbf{Potential Function Stability:}
The potential function training exhibited a similar trend, reducing loss by 99.82\%. While the loss curve exhibited oscillatory behavior typical of SGD with fixed learning rates, it stabilized around a consistent mean, confirming that the potential function successfully captured the joint interaction dynamics of the multi-agent system.

\subsection{Qualitative Analysis: Emergent Blocking}
To validate the interactive capabilities of our framework, we qualitatively analyze the ego vehicle's behavior in two close-proximity racing scenarios, highlighting its ability to proactively block approaching opponents. Figure \ref{fig:comparison} provides an overview of the two scenarios and compares the baseline planner \cite{ogretmen2024SamplingBasedb} to our approach.

\textbf{Scenario T1:} In the reactive configuration, the ego vehicle tracks the optimal raceline without incorporating the opponent's state into the lateral cost evaluation. Consequently, the opponent rapidly closes the longitudinal gap, resulting in a collision at the corner exit due to the speed differential. In contrast, our proposed planner generates an \ac{IRP} with a calculated lateral deviation to the left of the raceline in the braking zone. Guided by the modified cost function, the ego vehicle occupies the inside line. This induces a deceleration in the opponent's reactive planner, preventing the overtake and preserving the ego vehicle's leading position through the turn sequence.

\textbf{Scenario T5:} During the approach to the hairpin, the baseline planner maintains the nominal raceline, leaving the inner track geometry unobstructed and enabling the opponent to execute a successful inner overtake. When utilizing the hybrid planner, the online evaluation of the potential function yields interaction parameters that shift the \ac{IRP} toward the inner track boundary early in the approach phase. Tracking this strategic bias constrains the opponent to a dynamically inferior outer trajectory, allowing the ego vehicle to retain its track position and exit the corner with an increased longitudinal gap.

These qualitative results confirm that the gradient-based optimization of the interaction parameters effectively maps multi-agent game-theoretic evaluations into spatial trajectory biases, yielding dynamically feasible and competitively advantageous maneuvers.

\subsection{Quantitative Analysis}
To evaluate the statistical performance and robustness of the proposed framework beyond isolated scenarios, we analyzed the results of the large-scale randomized scenario set defined in Section \ref{sec:scenarios}. We define two primary metrics for this evaluation:
\begin{itemize}
    \item \textbf{Number of collisions:} The absolute number of episodes resulting in a collision involving the ego vehicle.
    \item \textbf{Net Position Change ($\Delta P$):} The average number of positions gained or lost by the end of the scenario. An overtake yields $+1$, while being overtaken yields $-1$.
\end{itemize}

The baseline reactive planner recorded $14$ collisions and achieved an average net position change of $\Delta P = 0.54$. In contrast, the proposed hybrid interaction-aware planner (configured with an optimal strategy weight from Section \ref{sec:impact_interaction_weight}) outperformed the baseline on both metrics. The number of collisions was reduced to $6$ incidents. Furthermore, the competitive performance improved, achieving an average net position change of $\Delta P = 0.61$. This indicates that the integration of the \ac{IRP} does not simply induce collision avoidance, but enables the ego vehicle to claim and defend positions more effectively than a standard sampling-based approach.

\subsection{Impact of the Interaction Weight}
\label{sec:impact_interaction_weight}
We now investigate the influence of the strategy weight $w_{\mathrm{strat}}$, i.e. the bias towards the \ac{IRP} in the cost function calculation, on the overall performance. 
To analyze this trade-off, we vary $w_{\mathrm{strat}}$ logarithmically from $0$ to $10^8$. Table \ref{tab:ablation} presents the results from this study.

\begin{table}[htbp]
    \centering
    \captionsetup{font=footnotesize, textfont=sc, labelfont=sc, labelsep=newline, justification=centering}
    \caption{Interaction weight $w_{\mathrm{strat}}$ influence over 200 runs}
    \label{tab:ablation}
    \small 
    \begin{tabular}{lcc}
        \toprule
        \textbf{Strategy Weight $w_{\mathrm{strat}}$} & \textbf{Collisions} & \textbf{Avg. $\Delta P$} \\
        \midrule
        $0$ (Baseline \cite{ogretmen2024SamplingBasedb}) & $14$ & $0.54$ \\
        $10^2$ & $11$ & $0.56$ \\
        $\mathbf{10^4}$ & $\mathbf{6}$ & $\mathbf{0.61}$ \\
        $10^6$ & $9$ & $0.48$ \\
        $10^8$ & $21$ & $0.32$ \\
        \bottomrule
    \end{tabular}
\end{table}

The framework's performance exhibits a high sensitivity to the tuning of this parameter:
\begin{itemize}
    \item \textbf{Low Strategy Cost ($w_{\mathrm{strat}} = 10^2$):} Trajectory selection is strictly dominated by the baseline performance cost $J_{\mathrm{perf}}$. The lateral bias introduced by the \ac{IRP} is mathematically insufficient to overcome the penalties associated with jerk and raceline deviation. Consequently, the generated trajectories fail to maintain the necessary lateral displacement for effective blocking, yielding a suboptimal competitive standing.
    \item \textbf{Balanced Configuration ($w_{\mathrm{strat}} = 10^4$):} This parameterization yields the highest net position change ($\Delta P = 0.61$). The geometric bias of the \ac{IRP} is adequately scaled to induce strategic lateral shifts, while the baseline cost $J_{\mathrm{perf}}$ retains sufficient mathematical influence to ensure dynamic feasibility and collision avoidance during close-proximity multi-agent interactions.
    \item \textbf{Over-Aggressive Tracking ($w_{\mathrm{strat}} \geq 10^6$):} The strategy cost $J_{\mathrm{strat}}$ overshadows the baseline performance metrics, effectively enforcing the \ac{IRP} tracking as a hard constraint. Because the \ac{IRP} is a purely geometric projection that does not fully account for complex, high-speed multi-agent kinematics, strict spatial adherence generates dynamically suboptimal trajectories. Trajectory selection heavily penalizes necessary evasive deviations from the \ac{IRP}, which directly correlates with a spike in collision rates and a severe degradation of longitudinal progress.
\end{itemize}

\subsection{Runtime Performance}
To ensure the feasibility of the online gradient ascent within high-frequency planning loops, we analyzed the computational overhead of the game-theoretic module. The strategy parameter optimization typically stabilizes within the first 5 to 10 iterations. Consequently, the total inference time required to optimize $\theta$ and generate the complete \ac{IRP} consistently remained between 10 ms and 17 ms across all experimental runs. Considering that standard local motion planners in autonomous racing operate at frequencies between 10 Hz and 20 Hz, this low latency demonstrates that the interaction module is suitable for online deployment.

\section{Conclusion}
This paper presented a hybrid interaction-aware motion planning framework for autonomous racing, bridging the gap between game-theoretic reasoning and sampling-based trajectory generation. By integrating an online-optimized \ac{IRP} as a dynamic cost bias, we enabled a standard sampling planner to exhibit strategic behaviors previously reserved for computationally intensive optimization-based methods.

Our simulation results on the Yas Marina Circuit demonstrate that the proposed architecture successfully generates emergent defensive maneuvers, such as blocking, allowing the ego vehicle to maintain its competitive position against aggressive opponents. We further identified a trade-off between IRP tracking aggressiveness and overall competitive performance, showing that a balanced tuning of the strategy cost weight is essential to retain collision-free maneuverability while actively defending positions.

Future work will focus on two main directions. First, we aim to validate the framework on a real-world autonomous race vehicle to assess robustness against perception noise and model mismatches. Second, investigate developing an adaptive, situation-aware scheduling mechanism for the strategy weight $w_{\mathrm{strat}}$ to dynamically balance strategic aggressiveness with kinematic feasibility based on local interaction complexity.

%% file: figs/teaser_fig.tex
\begin{figure}[!t]
    \centering
    
    \captionsetup[subfigure]{font=footnotesize, width=0.95\linewidth}

    \begin{subfigure}{\linewidth}
        \centering
        \begin{tikzpicture}[font=\scriptsize]
            
            \node[inner sep=0pt] at (0.5,0) {\includegraphics[height=2.5mm]{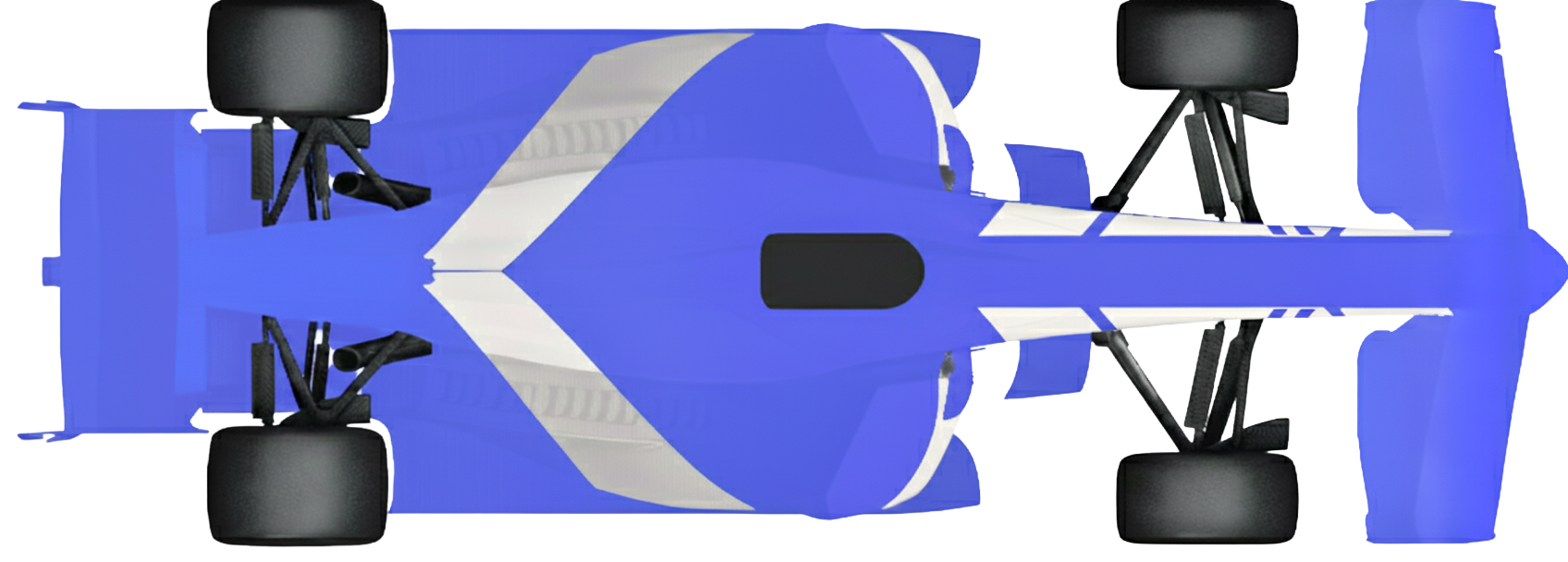}};
            \node[align=left, anchor=west] at (0.8,0) {ego \\ vehicle};
        
            \node[inner sep=0pt] at (2.3,0) {\includegraphics[height=2.5mm]{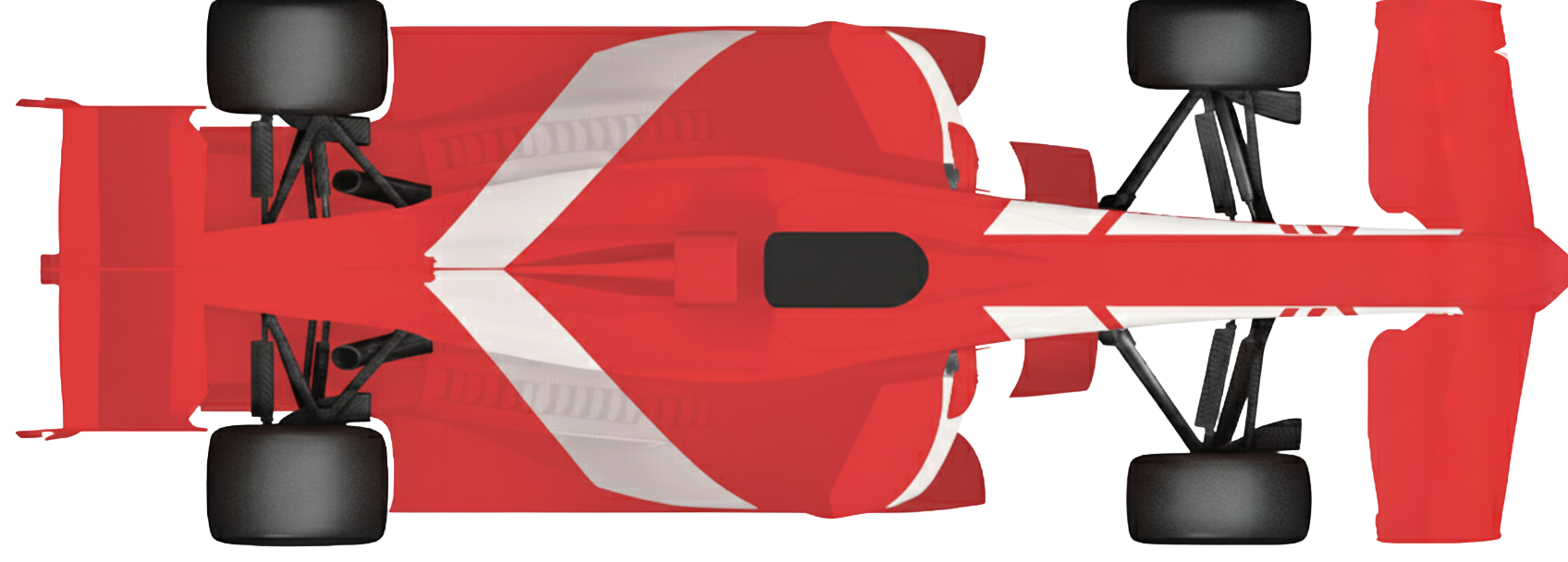}};
            \node[align=left, anchor=west] at (2.6,0) {opponent};
    
            \draw[thick, Blue, line width=1.5pt] (3.9,0.1) -- (4.4,0.1);
            \draw[thick, FerrariRed, line width=1.5pt] (3.9,-0.1) -- (4.4,-0.1);
            \node[align=left, anchor=west] at (4.5, 0) {driven\\trajectories};
    
            \draw[thick, IRPBlue, line width=1.5pt] (6.0,0.0) -- (6.5,0.0);
            \node[align=left, anchor=west] at (6.6, 0) {interaction\\reference path};

            \node[anchor=north west, inner sep=0pt] at (0,-0.4) {\fbox{\includegraphics[width=0.95\linewidth, trim={3.0cm 0.5cm 1.5cm 1.5cm},clip]{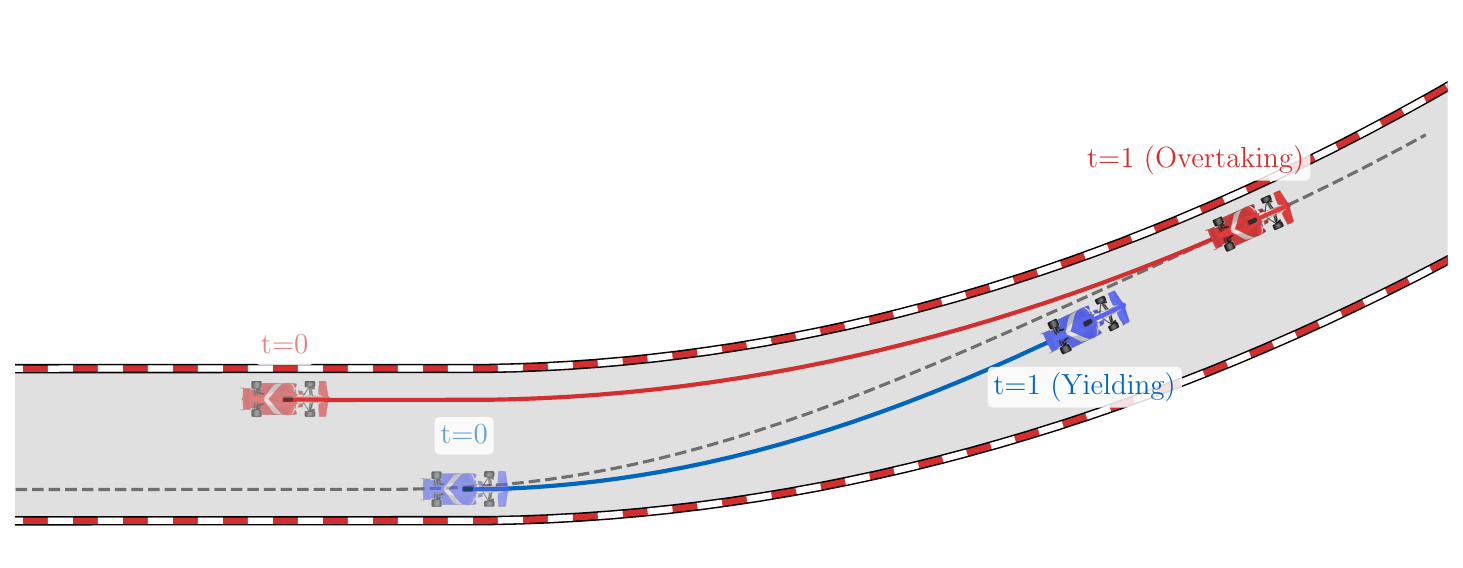}}};
        \end{tikzpicture}
        \caption{Reactive sampling-based planner: The ego vehicle yields to the approaching opponent to avoid a collision.}
        \label{fig:RL_intro_a}
    \end{subfigure}
    
    \vspace{0.3cm} 
    
    \begin{subfigure}{\linewidth}
        \centering
        \fbox{\includegraphics[width=0.95\linewidth, trim={3.0cm 0.5cm 1.5cm 1.5cm},clip]{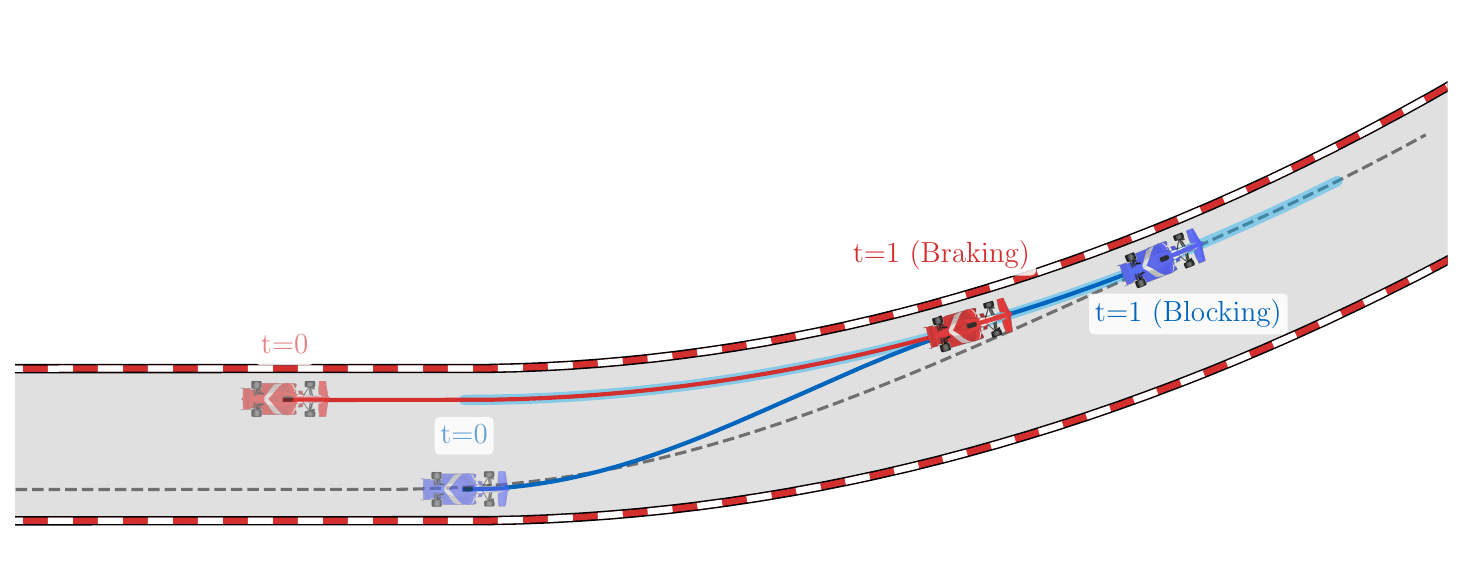}}
        \caption{Proposed game theory-informed sampling-based planner: The ego vehicle actively blocks the opponent to prevent an overtake.}
        \label{fig:RL_intro_b}
    \end{subfigure}

    \caption{Comparison of our proposed planner (b) against a reactive sampling-based planner baseline (a). The dashed gray line indicates the optimal raceline.}
    \label{fig:RL_intro}
\end{figure}